\def\year{2018}\relax
\documentclass[letterpaper]{article} %DO NOT CHANGE THIS
\usepackage{aaai18}  %Required
\usepackage{times}  %Required
\usepackage{helvet}  %Required
\usepackage{courier}  %Required
\usepackage{url}  %Required
\usepackage{graphicx}  %Required
\usepackage[utf8]{inputenc} % allow utf-8 input
\usepackage[T1]{fontenc}    % use 8-bit T1 fonts
\usepackage{url}            % simple URL typesetting
\usepackage{booktabs}       % professional-quality tables
\usepackage{amsfonts}       % blackboard math symbols
\usepackage{nicefrac}       % compact symbols for 1/2, etc.
\usepackage{microtype}      % microtypography

\usepackage{amsmath}
\usepackage{amssymb}
\usepackage{esint}
\usepackage{times}
\usepackage{epsfig}
\usepackage{epstopdf}
\usepackage{caption}
\usepackage{subcaption} 
\usepackage{booktabs}
\usepackage{color}
\usepackage[export]{adjustbox}
\usepackage{wrapfig}
\usepackage{lipsum}
\usepackage[capbesideposition=right]{floatrow}
\usepackage{xspace}

\newcommand{\myparagraph}[1]{\vspace{.5em}\noindent\textbf{#1}}
\newcommand{\bx}{\mathbf{x}\xspace}
\newcommand{\bw}{\mathbf{w}\xspace}

\newcommand{\btheta}{\boldsymbol{\theta}\xspace}
\newcommand{\balpha}{\boldsymbol{\alpha}\xspace}
\newcommand{\brho}{\boldsymbol{\rho}\xspace}
\newcommand{\calY}{\mathcal{Y}\xspace}
\newcommand{\D}{\mathcal{D}\xspace}

\newcommand{\iid}{\emph{i.i.d.}\xspace}
\newcommand{\ie}{\emph{i.e.}\xspace}
\newcommand{\eg}{\emph{e.g.}\xspace}
\newcommand{\cf}{\emph{cf.}\xspace}
\newcommand{\wrt}{\emph{w.r.t.}\xspace}
\newcommand{\Eq}{Eq.\xspace}
\newcommand{\Fig}{Fig.\xspace}

\newcommand{\etal}{\emph{et al.}\xspace}
\begin{document}
\newcommand{\UOA}{$^1$}
\newcommand{\AMZ}{$^2$}
\title{Joint Learning of Set Cardinality and State Distribution}
\author{S. Hamid Rezatofighi\UOA \quad 
Anton Milan\AMZ\thanks{The work was carried out at the University of Adelaide.} \quad
Qinfeng Shi\UOA \quad
Anthony Dick\UOA \quad
Ian Reid\UOA \\
	\\
	\UOA School of Computer Science, The University of Adelaide, Australia\\
	\AMZ Amazon Development Center, Germany\\
	{firstname.lastname}@adelaide.edu.au
	\\
}
\maketitle

\begin{abstract}
We present a novel approach for learning to predict sets using deep learning. In recent years, deep neural networks have shown remarkable results in computer vision, natural language processing and other related problems. Despite their success, traditional architectures suffer from a serious limitation in that they are built to deal with structured input and output data, \ie vectors or matrices. Many real-world problems, however, are naturally described as sets, rather than vectors. Existing techniques that allow for sequential data, such as recurrent neural networks, typically heavily depend on the input and output order and do not guarantee a valid solution. Here, we derive in a principled way, a mathematical formulation for set prediction where the output is permutation invariant.
% \fixmeh{"which is permutation invariant", We can remove this. The sentence is actually correct and the output is permutation invariant. But it is misleading about learning to be permutation invariant }. 
In particular, our approach jointly learns both the cardinality and the state distribution of the target set. We demonstrate the validity of our method on the task of multi-label image classification and achieve a new state of the art on the PASCAL VOC and MS COCO datasets. 
\end{abstract}

\section{Introduction}

\noindent Recently, deep structured networks such as deep convolutional (CNN) and recurrent (RNN) neural networks
have become increasingly popular in artificial intelligence, showing remarkable performance on many real-world problems, including scene classification~\cite{Krizhevsky:2012:NIPS}, speech recognition~\cite{hinton2012deep}, gaming~\cite{mnih2013playing,mnih2015human}, semantic segmentation~\cite{Papandreou:2015:ICCV}, and image captioning~\cite{Johnson:2016:CVPR}. However, like most machine learning techniques, current deep learning approaches are based on conventional statistics and require the problem to be formulated in a structured way. In particular, they are designed to learn a model for a distribution (or a function) that maps a structured input, typically a vector, a matrix, or a tensor, to a structured output.

Consider the task of image classification as an example. The goal here is to predict a label (or a category) of a given image. The most successful approaches address this task with CNNs, \ie by applying a series of convolutional layers followed by a number of fully connected layers~\cite{Krizhevsky:2012:NIPS,Simonyan:2014:VGG,Szegedy:2014:Inception,He:2016:ResNet}. The final output layer is a fixed-sized vector with the length corresponding to the number of categories in the dataset (\eg~1000 in the case of the ILSVR Challenge~\cite{Russakovsky:2015:ILSVRC}). Each element in this vector is a score or probability for one particular category such that the final prediction corresponds to a probability distribution over all classes. The difficulty arises when the number of classes is unknown in advance and in particular varies for each example. Then, the final output is generated heuristically by a discretization process such as choosing the $k$ highest scores~\cite{gong2013deep,Wang_2016_CVPR}, which is not part of the learning process. 
This shortcoming concerns not only image tagging but also other problems like detection or graph optimization, where connectivity and graph size can be arbitrary.

We argue that such problems can be naturally expressed with sets rather than vectors. As opposed to a vector, the size of a set is not fixed in advance, and it is invariant to the
ordering of entities within it. Therefore, learning approaches built on conventional statistics cannot
be the right choice for these problems. In this paper, we propose a learning approach based on point processes and finite set statistics to deal with sets in a principled manner. More specifically, in the presented model, we assume that the input (the observation) is still  structured, but the output is modelled as a set. Our approach is inspired by a recent work on set learning using deep neural networks~\cite{rezatofighi2017deepsetnet}. The main limitation of that work, however,  is that the approach employs two sets of independent weights (two independent networks) to generate the cardinality and state distributions of the output set. In addition, to generate the final output as a set, sequential inference has to be applied instead of joint inference. In this paper, we derive a principled formulation for performing both learning and inference steps jointly. The main contribution of the paper is summarised as follows:
%\vspace{-1em}
\begin{enumerate}
	\setlength{\itemsep}{1pt}
	\setlength{\parskip}{0pt}
	\setlength{\parsep}{0pt}
	\item We present a novel way to learn both cardinality and state distributions jointly within a single deep network. Our model is learned end-to-end to generate the output set. 
	\item We perform the inference step both jointly and optimally. We show how we can generate the most likely (the optimal) set using MAP inference for our given model.    	
	\item Our approach outperforms existing solutions and achieves state-of-the-art results on the task of multi-label image classification on two standard datasets. 
\end{enumerate}

\section{Related Work}
\label{related work}
Handling unstructured input and output data, such as sets or point patterns, for both learning and inference is an emerging field of study that has generated substantial interest in recent years. Approaches such as mixture models~\cite{blei2003latent,hannah2011dirichlet,tran2016clustering}, learning distributions from a set of samples~\cite{muandet2012learning,oliva2013distribution}, model-based multiple instance learning~\cite{vo2017model} and novelty detection from point pattern data~\cite{vo2016model}, can be counted as few out many examples that use point patterns or sets as input or output and directly or indirectly model the set distributions. However, existing approaches often rely on parametric models, \eg~the elements in output sets needs to be derived from
an independent and identically distributed (\iid) Poisson point process distribution~\cite{adams2009tractable,vo2016model}.  Recently, deep learning has enabled us to use less parametric models to capture highly complex mapping distributions between structured inputs and outputs. Somewhat surprisingly, there are only few works on learning sets using deep neural networks. One interesting exception in this direction
is the recent work of Vinyals \etal~\shortcite{vinyals2015order}, which uses an RNN to read and predict sets. However, the output is still assumed to have an ordered structure, which contradicts the orderless (or permutation invariant) property of sets. Moreover, the framework can be used in combination with RNNs only and cannot be trivially extended to any arbitrary learning framework such as feed-forward architectures. Another recent work proposed by Zaheer~\etal~\shortcite{zaheer2017deep} is a deep learning framework which can deal with sets as input with different sizes and permutations. However, the outputs are either assumed to be structured, \eg~a scalar as a regressing score, or a set with the same entities of the input set, which prevents this approach to be used for the problems that require output sets with arbitrary entities. Perhaps the most related work to our problem is a deep set network recently proposed by Rezatofighi~\etal~\shortcite{rezatofighi2017deepsetnet} which seamlessly integrates a deep learning framework into set learning in order to learn to predict sets in 
two challenging computer vision applications, image tagging and pedestrian detection. However, the approach requires to train two independent networks to model a set, one for cardinality and one for state distribution. Our approach is largely inspired by this latter work but overcomes its limitation on independent learning and inference.

To validate our model, we apply it on the multi-label image classification task. Despite its relevance, there exists rather little work on this problem that makes use of deep architectures. One example is Gong~\etal~\shortcite{Gong:2013:arxiv}, who combine deep CNNs with a top-$k$ approximate ranking loss to predict multiple labels. Wei~\etal~\shortcite{Wei:2014:arxiv} propose a Hypotheses-Pooling architecture that is specifically designed to handle multi-label output. While both methods open a promising direction, their underlying architectures largely ignore the correlation between multiple labels. To address this limitation, recently, Wang~\etal~\shortcite{Wang_2016_CVPR} proposed a model that combines CNNs and RNNs to predict an arbitrary number of classes in a sequential manner. 
RNNs, however, are not suitable for set prediction mainly for two reasons. 
First, the output represents a sequence and not a set, and is thus highly dependent on the prediction order, as was shown recently by Vinyals~\etal~\shortcite{vinyals2015order}. 
Second, the final prediction may not result in a feasible solution (\eg it may contain the same element multiple times), such that post-processing or heuristics such as beam search must be employed~\cite{Vinyals:2015:NIPS,Wang_2016_CVPR}. 
Here we show that our approach not only guarantees to always predict a valid set, but also outperforms previous methods. 

\section{Background }
\label{background}

To better explain our approach, we first review some mathematical background and introduce the notation used throughout the paper.
In statistics, a continuous random variable $y$ is a variable that can take an infinite number of possible values. A continuous random vector can be defined by stacking several continuous random variables into a fixed length vector, $Y=\left(y_1,\cdots,y_m\right)$. The mathematical function describing the possible values of a continuous random vector and their associated joint probabilities is known as a probability density function (PDF) $p(Y)$ such that
$\int p(Y)dY = 1.$
%**********************************************************************************************
% \begin{equation}\label{eq: pdf}
% \begin{aligned}
% %\int_{-\infty}^{\infty} p(Y)dY = 1.
% \int p(Y)dY = 1.
% \end{aligned}
% \end{equation}
%**********************************************************************************************

In contrast, a random finite set (RFS) $\calY$ is a finite-set valued random variable $\calY=\left\{y_1,\cdots,y_m\right\}$. The main difference between an RFS and a random vector is that for the former, the number of constituent variables, $m$, is random and the variables themselves are random and unordered.
Throughout the paper, we use $\calY$ for a set with unknown cardinality, $\calY^m$ for a set with known cardinality $m$ and $Y=\left(y_1,\cdots,y_m\right)$ for a vector (or an ordered set) with known dimension $m$.

A statistical function describing a finite-set variable $\calY$ is a
combinatorial probability density function $p(\calY)$ which consists of a discrete probability distribution, the so-called cardinality distribution, and a family of joint probability densities on both the number and the values of the constituent variables~\cite{mahler2007statistical,vo2017model}, \ie
\begin{equation}
\begin{aligned}
p(\calY) & = p(m)U^m p_m(\{y_{1},y_{2},\cdots,y_{m}\})\\
& = p(m)m!U^m p_m(y_{1},y_{2},\cdots,y_{m}),
\end{aligned}
\label{eq:setprob}
\end{equation} 
where $p(m)$ is the cardinality distribution of the set $\calY$ and  $p_m(\{y_{1},y_{2},\cdots,y_{m}\})$ is a symmetric joint probability density distribution of the set $\calY^m$ given known cardinality $m$. The normalisation factor $m!=\prod_{k=1}^m k$ between $p_m(\calY^m)$ and $p_m(Y)$ appears because the probability density for a set with known cardinality $\calY^m$ must be equally distributed among all the $m!$ possible permutations of the corresponding vector $Y$~\cite{mahler2007statistical,vo2017model}. $U$ is the unit of hyper-volume in the feature space, which
cancels out the unit of the probability density
$p_m(\cdot)$ making it unit-less, and thereby avoids the unit
mismatch across the different dimensions (cardinalities). Without this normalizing constant, the comparison between probabilities of the sets with different cardinalities is not properly defined because a distribution with the smallest set size will always have the highest probabilities. For example, $p(y_1)\geq p(y_1,y_2)$ always holds regardless of the particular choice for $y_1$ and $y_2$. Please refer to~\cite{vo2017model} for an intuitive discussion.

Finite Set Statistics
provides powerful and practical mathematical tools for dealing with random finite sets, based on the notion
of integration and density that is consistent with the point process theory~\cite{mahler2007statistical}.\footnote{A random finite set can be viewed as a simple finite point process~\cite{baddeley2007spatial}.} For example, similar to the definition of a PDF for a random variable, the PDF of an RFS must sum to unity over all possible cardinality values and all possible element values as well as their permutations. This type of statistics, which is derived from the point process stochastic process, defines basic
mathematical operations on finite sets such as functions, derivatives and integrations as well as other statistical
tools such as probability density function of a random finite set and its statistical moments~\cite{mahler2007statistical,vo2017model}.   For
further details on point processes, we refer the reader to
textbooks such as~\cite{chiu2013stochastic,daley2007introduction,moller2003statistical}.

Conventional machine learning approaches, such as Bayesian learning and convolutional neural networks, have been proposed to learn the optimal parameters $\bw^*$ of the distribution $p(Y|\bx,\bw^*)$ which maps the input vector $\bx$ to the \emph{output vector} $Y$.
In this paper, we instead propose an approach that can learn the parameters $\bw^*$ for a set distribution that allows one to map the input vector $\bx$ to the \emph{output set} $\calY$, \ie  $p(\calY|\bx,\bw^*)$. For mathematical convenience, we use an \iid-cluster point process model. Moreover, we target applications where the order of the outputs during training is irrelevant, \eg multi-label image classification. Modifying the application or
the \iid assumption to non-\iid set elements, may require to deal with the complexity of permutation invariant property of sets during the learning step, which leads to serious
mathematical complexities and is left for future work.

\section{Joint Deep Set Network}
\label{JDSN}

We follow the convention introduced in~\cite{rezatofighi2017deepsetnet} and define a training set $\D = \{(\bx_{i},\calY_{i})\}$,
% \begin{eqnarray*}
% 	\calY & = & \{y_{1},y_{2},\ldots,y_{m}\}\qquad y_{k}\in\mathbb{R}^{d}, \forall k\\ % \text{or}\quad\mathbb{Z}^{d} \\
% 	\D & = & \{\calY_{i},\bx_{i}\}\qquad\qquad\quad\quad\forall\bx_{i}\in\mathbb{R}^{l}\quad\forall i=1,\ldots,n,
% \end{eqnarray*}
where each training sample $i=1,\ldots,n$ is a pair consisting of an input feature (\eg image), $\bx_{i}\in\mathbb{R}^{l}$ and an output (or label) set
$\calY_{i} = \{y_{1},y_{2},\ldots,y_{m_i}\}, y_{k}\in\mathbb{R}^{d}, m_i\in\mathbb{N}^0 $. In the following we will drop the instance index $i$ for better readability. Note that $m:=|\calY|$ denotes the cardinality of set $\calY$.
Following the definition in Eq.(~\ref{eq:setprob}), the probability density of a set $\calY$ with an unknown cardinality is defined as
\begin{equation}
\begin{aligned}
p(\calY|\bx,\bw) =& p(m|\bx,\bw)\times U^m \\ &\times p_m(\{y_{1},y_{2},\cdots,y_{m}\}|\bx,\bw),
\end{aligned}
\end{equation} 
where  $\bw$ denotes the collection of parameters which model both the \emph{cardinality} distribution of the set elements $p(m|\cdot)$ as well as the parameters of $y_{k}$ that model the joint distribution of set element \emph{values} for a fixed cardinality $p_m(\{y_{1},y_{2},\cdots,y_{m}\}|\cdot)$. Note that in contrast to previous works~\cite{rezatofighi2017deepsetnet,vo2016model,vo2017model} that assume that two sets of independent parameters (two independent networks) are required to represent 
the set distribution $p(\calY|\cdot)$, we will show that one set of parameters $\bw$ is sufficient to learn this distribution and as it turns out also yields better performance.

The above formulation represents the probability density
of a set which is very general and completely independent
of the choices of both cardinality and state distributions.
It is thus straightforward to transfer it to many applications
that require the output to be a set. However, to
make the problem amenable to mathematical derivation and
implementation, we adopt two assumptions: \emph{i)} the outputs (or labels) in the set are derived from an independent
and identically distributed (\iid)-cluster point process model, and \emph{ii)} their cardinality follows
a categorical distribution parameterised by event probabilities $\brho$.
% \fixmea{ok, I still find it a bit strange that you say parameter $\brho$ and two lines later $\brho$ is the vector of probabilities}\fixmeh{The parameters for a categorical distribution called the event probabilities and $\brho$ is the parameter for the categorical distribution. still I don't know what is confusing here for you}
% ok, didnt know the parameter was called scent probabilities
Thus, we can write the distribution
as
\begin{equation}
\begin{aligned}
p(\calY|\bw,\bx) = \int p(m|\brho)&p(\brho|\bx,\bw) d\brho \times  U^m \\&\times  \left(\prod_{y\in\calY^m}p(\{y\}|\bx,\bw)\right),
\label{eq:posterior_general} 
\end{aligned}
\end{equation} 
where $p(\{y\}|\cdot,\cdot)$ denotes the probability of taking on the state $y$ in a singleton set $\{y\}$, and $\brho = (\rho_1,\ldots,\rho_M)$ is the vector of event probabilities, 
\emph{i.e.} $\sum_{i=1}^M \rho_i = 1$ and $\rho_i>0,\forall i\in\{1,\ldots,M\}$.

\subsection{Posterior distribution}
\label{sec:posterior}
To learn the parameters $\bw$, we assume that the training samples are independent from each other and that the distribution $p(\bx)$ from which the input data is sampled is independent from both the output and the parameters. 
Then, the posterior distribution over the parameters can be derived as
\begin{equation}
\begin{aligned}
p(\bw|\D) &\propto p(\D|\bw)p(\bw)\\
&=\prod_{i=1}^{n}\left[\int p(m_{i}|\brho)p(\brho|\bx_{i},\bw)d\brho\times 
U^{m_i}\right.\\&\quad\quad\times \left.\left(\prod_{y\in\calY^{m_i}_i}p(\{y\}|\bx_i,\bw)\right) \right]p(\bw).
\end{aligned} 
\label{eq:posterior}
\end{equation}
%where $K$ is a normalizer defined as 
%\begin{equation*}
%K =  \int\prod_{i=1}^{n} \left[\int p(m_{i}|\rho)p(\rho|\bx_{i},\bw)d\rho\times 
%m_{i}!\times U^{m_i}\times \left(\prod_{y\in\calY_i}p(y|\bx_i,\bw)\right) p(\bx_{i})\right]p(\bw)\quad d\bw.
%\end{equation*}
%The probability
%$p(\bx_{i})$ can be eliminated as it appears in both the numerator and the denominator. Therefore,  
%%
%\begin{equation}
%p(\bw|\D) = \frac{1}{\tilde{K}}\prod_{i=1}^{n}\left[\int p(m_{i}|\rho)p(\rho|\bx_{i},\bw)d\rho\times 
%m_{i}!\times U^{m_i}\times \left(\prod_{y\in\calY_i}p(y|\bx_i,\bw)\right)\right]p(\bw),
%\label{eq:posterior_m}
%\end{equation}
%where 
%\begin{equation*}
%\tilde{K} = \int \int\prod_{i=1}^{n} \left[\int p(m_{i}|\rho)p(\rho|\bx_{i},\bw)d\rho\times 
%m_{i}!\times U^{m_i}\times \left(\prod_{y\in\calY_i}p(y|\bx_i,\bw)\right)\right]p(\bw)\quad d\bw.
%\end{equation*}

\begin{figure*}[t]
	\begin{minipage}[b]{.48\linewidth}
		\begin{center}
			\includegraphics[trim={11.5cm 7.2cm 12cm 6cm},clip,width=0.7\linewidth]{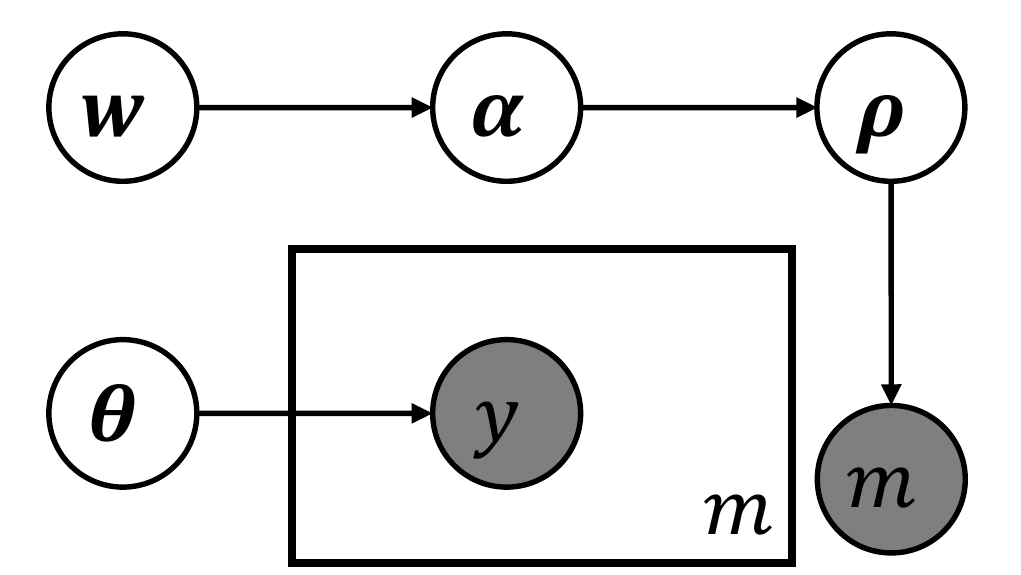}
			\centerline{(a)}\medskip
		\end{center}
	\end{minipage}
	\hfill
	\begin{minipage}[b]{.48\linewidth}
		\begin{center}
			\includegraphics[trim={12cm 7cm 12cm 6.5cm},clip,width=0.7\linewidth]{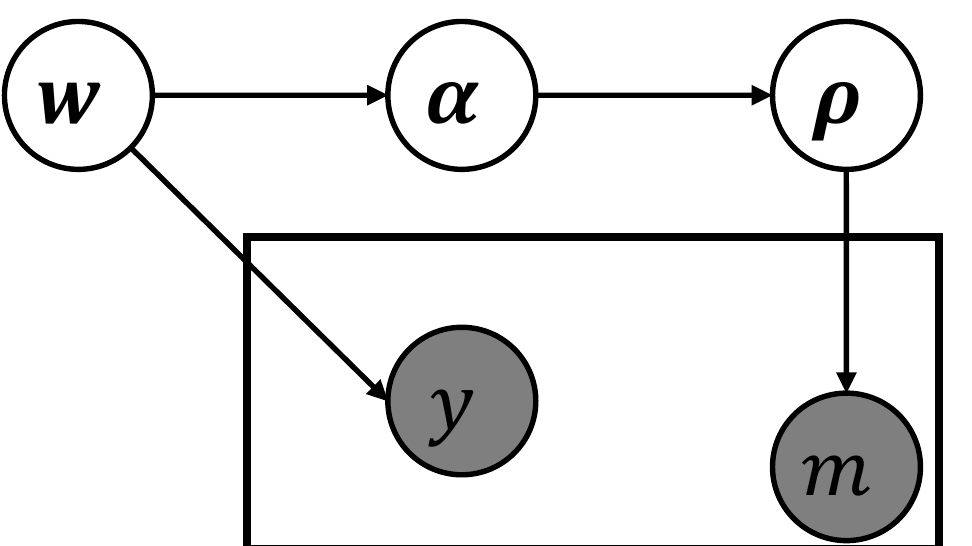}
			\centerline{(b)}\medskip
		\end{center}
	\end{minipage}
	\caption{Comparison of the graphical models: a) The set learning approach introduced in ~\cite{rezatofighi2017deepsetnet} by replacing  Dirichlet-Categorical as its cardinality distribution; (b) our proposed joint set learning. The work in ~\cite{rezatofighi2017deepsetnet} first  predicts the cardinality $m$, and then the labels $y$s given $m$. There is a separation between parameters $\bw$ and $\btheta$. Consequently, an incorrect $m$ predicted via $\bw$ can not be fixed by $\btheta$. Our method only uses one joint parameter $\bw$ which aims to learn and predict the best $m$ and $y$'s \emph{jointly}. $y$ and $m$ are shaded as they are observed in the training data. Note that $m$ is a variable in our model (b) which determines the repetition of the plates (\ie the number of $y$'s). Removing the top-right chain in (a) recovers the  traditional vector based (non-set based) method. }

%	\vspace{-1.0em}
	\label{fig:pgm}
\end{figure*}

A closed-form solution for the integral in Eq.~(\ref{eq:posterior}) can be obtained by using conjugate priors:
\begin{eqnarray*}
	m & \sim & \text{Cat}(m;\brho)\\
	\brho & \sim & \text{Dir}(\brho;\balpha(\bx,\bw))\\
	&&\balpha(\bx,\bw)>0\quad\forall\bx,\bw,
\end{eqnarray*}
where $\text{Cat}(\cdot,\brho)$ and $\text{Dir}(\cdot;\balpha)$ represent respectively a categorical distribution with the event probabilities $\brho = (\rho_1,\ldots,\rho_M)$ and a Dirichlet distribution with parameters $\balpha = (\alpha_1,\ldots,\alpha_M)$. Moreover, $p(\bw)$ can be assumed a zero-mean normal distribution with covariance equal to $\sigma^{2}\mathbf{I}$, \ie $p(\bw) = \mathcal{N}(\cdot;0,\sigma^{2}\mathbf{I})$. The key difference between our method and \cite{rezatofighi2017deepsetnet} is that we only need to use one network as opposed to two networks used in the previous work. It is important to note that our method \emph{jointly} predicts both cardinality and the set elements as opposed to sequentially predicting the cardinality first and then the set elements as previously done in \cite{rezatofighi2017deepsetnet}. We have provided a comparison between the graphical models of both methods in terms of plate notation in Fig.~\ref{fig:pgm} to further illustrate their differences.

We assume that the cardinality follows a categorical distribution whose event probabilities vector $\brho$ is estimated from a Dirichlet distribution with parameters $\balpha$,  which can be directly estimated from the input data $\bx$. Note that the cardinality distribution in Eq.~(\ref{eq:posterior_general}) can be replaced
by any other discrete distribution, \eg Poisson, binomial or negative binomial (\cf~\cite{rezatofighi2017deepsetnet}). 
Here, we use the categorical distribution as the cardinality model, which better suits the task at hand. The rationale here is that Poisson and negative binomial are long-tailed distributions and their variance increases with their mean. Therefore, the final model will have more uncertainty (and possibly a higher error) in estimating the cardinality of the sets with high values. In contrast,  the categorical distribution does not have the drawback of correlating its mean and variance. Moreover, in the image tagging application, the maximum cardinality is often known and there is no need to use long-tailed distributions, which are more suitable for the applications where the maximum cardinality is unknown.

Consequently, the integral in Eq.~(\ref{eq:posterior}) is simplified
and forms a Dirichlet-Categorical distribution
\begin{equation}
DC\left(m;\balpha\right)  =  \frac{\alpha_m+C_m}{\sum_{\acute{m}}\alpha_{\acute{m}}+C}\\, 
\label{eq:DC}
\end{equation}
where $C_m$ is the number of samples in the training set with cardinality $m$, and $C$ is the total number of training samples. Finally, the full posterior distribution can be written as
\begin{equation}
\begin{aligned}
p(\bw|\D)  \propto\prod_{i=1}^{n}&\bigg[DC\left(m_{i};\balpha(\bx_{i},\bw)\right)\times U^{m_i}\\&\times \left(\prod_{y\in\calY^{m_i}_i}p(\{y\}|\bx_i,\bw)\right)\bigg]p(\bw).
\label{eq:full-posterior}
\end{aligned}   
\end{equation}

%According to the ranking function defined in~\cite{}, we can replace the above distribution by a proper unitless distribution function independent of the unit of hyper-volume $U$ as
%\begin{equation}
%\begin{aligned}
%p(\bw|\D)  \propto\prod_{i=1}^{n}\bigg[DC\left(m_{i};\balpha(\bx_{i},\bw)\right)\times \left(\prod_{y\in\calY_i}\frac{p(y|\bx_i,\bw)}{\parallel p(y|\bx_i,\bw)\parallel_2^2}\right)\bigg]p(\bw),
%\label{eq:full-posterior-final}
%\end{aligned}   
%\end{equation}
%where $\parallel p(y|\cdot)\parallel_2^2
%=
%\int
%p^2(y|\cdot)
%dy$ is the squared L2-norm of $p(y|\cdot)$.

\subsection{Learning}
\label{sec:learning}

For simplicity, we use a point estimate for the posterior $p(\bw|\D)$,
\textit{i.e.} $p(\bw|\D) = \delta(\bw=\bw^{*}|\D)$, where $\bw^{*}$ is computed using the MAP estimator, \ie $\bw^{*} = \arg\max_{\bw}\enspace \log\left(p\left(\bw|\D\right)\right)$.
Therefore, we have  
\begin{equation}
\begin{aligned}
\bw^{*} & =  
\arg\max_{\bw}\enskip\sum_{i=1}^{n}\bigg[\sum_{y\in\calY^{m_i}_i}\bigg(\log\big(p(\{y\}|\bx_i,\bw)\big)\bigg)\\
&+m_{i}\log U+\log\left(DC\left(m_{i};\balpha(\bx_{i},\bw)\right)\right)\bigg]-\gamma\|\bw\|_2^2.
\label{eq:map_complete}
\end{aligned}
\end{equation}

$p(\{y\}|\bx_i,\bw)$ describes a neural network with coefficients $\bw$ learned to map the input $\bx_i$ to the output (label) $y$. This function represents the \emph{state distribution} of each set element over the state space. $\gamma$ is the regularisation parameter, proportional to the predefined covariance parameter $\sigma$. This parameter is also known as the weight decay parameter and is commonly used in training neural networks. 

For example, in the application to multi-label image classification, $y\in\calY^{m_i}_i\subseteq\{\ell_1,\ell_2,\cdots,\ell_M\}$ represents the existence of a specific label $\ell_j$ in the input image instance $\bx_i$ from all pre-defined $M$ labels. In this application, we can rewrite an equivalent  binary formulation for the above MAP problem as
\begin{equation}
\begin{aligned}
\bw^{*}=   &\arg\max_{\bw}\enskip\sum_{i=1}^{n}\bigg[\sum_{\ell=1}^M z_i^{\ell}\log p(z_i^{\ell}|\bx_i,\bw)+\sum_{\ell=1}^M z^{\ell}_i \log U\\
& +\log DC\left(\sum_{\ell=1}^M z^{\ell}_i ;\balpha(\bx_{i},\bw)\right)\bigg]-\gamma\|\bw\|_2^2\\
=&\arg\max_{\bw}\enskip\sum_{i=1}^{n}\bigg[\sum_{\ell=1}^M z_i^{\ell}\log p(z_i^{\ell}|\bx_i,\bw)+
\\&\log DC\left(\sum_{\ell=1}^M z^{\ell}_i ;\balpha(\bx_{i},\bw)\right)\bigg]-\gamma\|\bw\|_2^2,
\end{aligned}
\label{eq:map_dual}
\end{equation}
where $z_i^{\ell}\in\{0,1\}$ represents the existence or non-existence of any specific label in the image $\bx_i$.  $p(z_i^{\ell}=1|\bx_i,\bw)$ can be defined as a binary logistic regression function  
\begin{equation*}
p(z_i^{\ell}=1|\bx_i,\bw) = \frac{\exp{O^\ell(x_i,\bw)}}{1+\exp{O^\ell(x_i,\bw)}},
\label{eq:BCE_loss}
\end{equation*}
where $O^{\ell}(x_i,\bw)$ is the network's predicted output corresponding to the $\ell^\text{th}$ label. 

Note that $\bw$ can
generally be learned using a number of existing machine learning techniques. In this paper we rely on deep CNNs to perform this task. More formally, to estimate $\bw^{*}$, we
compute the partial derivatives of the objective function in Eq.~(\ref{eq:map_dual}) and use standard backpropagation
to learn the parameters of the deep neural network.

\subsection{Inference}
\label{sec:inference}
Having learned the network parameters $\bw^{*}$, for a test image $\bx^{+}$, we use a MAP estimate to generate a set output as 
\begin{equation}
\calY^{*} 
= \arg\max_{\calY}\enspace p(\calY|\D,\bx^{+},\bw^{*}),
\end{equation}
where $p(\calY|\D,\bx^{+},\bw^{*})  \propto \int p(\calY|\bw,\bx^{+})p(\bw|\D) d\bw$, and $p(\bw|\D) = \delta(\bw=\bw^{*}|\D)$ as above. Therefore, the MAP estimate can be written as follows, 
\begin{equation}
\begin{aligned}
\calY^{*} 
& = \arg\max_{\calY}\enspace p(\calY|\D,\bx^{+},\bw^{*})\\
& = \arg\max_{\calY}\enspace\log\left(p(\calY|\D,\bx^{+},\bw^{*})\right)\\
& = \arg\max_{m,\calY^m} \enspace \log DC\left(m;\balpha(\bx^{+},\bw^{*})\right)+ m\log U\\
&\quad\quad\quad\quad\quad + \sum_{y\in\calY^m}\log\left(p(\{y\}|\bx^{+},\bw^{*})\right).
\end{aligned}
\label{eq:inference}
\end{equation}
Since the unit of hyper-volume $U$ in this application is unknown, we assume it as a constant hyper-parameter, estimated from the validation set of the data. 

To solve the above inference problem, we define the binary variable $z^{\ell}\in\{0,1\}$ for existence of each label similar to the learning process. Therefore, an equivalent formulation for Eq.~(\ref{eq:inference}) is
\begin{equation}
\begin{aligned}
&Z^{*}  =  \arg\max_{Z} \enskip\log DC\left(\sum_{\ell=1}^{M} z^{\ell};\balpha(\bx^{+},\bw^{*})\right)+\\&\sum_{\ell=1}^{M} z^{\ell}\log U+\sum_{\ell=1}^{M} z^{\ell}\log\left(\frac{\exp{O^\ell(\bx^{+},\bw^{*})}}{1+\exp{O^\ell(\bx^{+},\bw^{*})}}\right),
\end{aligned}
\label{eq:inference_final}
\end{equation}
where 
%\begin{equation}
$
Z=(z^{1},\cdots,z^{M})\in\{0,1\}^M. 
$
%\end{equation}
The above problem can be seen as  a combination of a higher-order term,  
\begin{equation}
f(\mathbf{1}^TZ,\balpha(\cdot)) = \log DC\left(\sum_{\ell=1}^{M} z^{\ell};\balpha(\cdot)\right), 
\end{equation}
which accounts for the total number of selected variables,  and a linear binary program, $C^TZ$,   where $C = (c^1,\cdots,c^M)$ and 
\begin{equation}
c^\ell = \log U +\log\left(\frac{\exp{O^\ell(\bx^{+},\bw^{*})}}{1+\exp{O^\ell(\bx^{+},\bw^{*})}}\right).
\end{equation}
Therefore, we can re-write it as
\begin{equation}
Z^{*} = \arg\max_{Z}\enspace f(\mathbf{1}^TZ,\balpha(\cdot))+C^TZ.
\end{equation}
% \label{eq:inference_final2}

Since for each specific cardinality $m=\mathbf{1}^TZ$, the most likely set corresponds to the $m$ highest values of $C$, the optimal solution for $m$ can be found efficiently when the sorted values of $C$, here denoted by $C_{st}=(c_{st}^1,\cdots,c_{st}^M)$, and $f(\cdot)$ is maximised \wrt $m$:
\begin{equation}
m^{*} = \arg\max_{m} \enspace f(m,\balpha(\cdot))+\sum_{\ell=1}^m c^{\ell}_{st}.
\label{eq:inference_equival}
\end{equation}
Then, the optimal $Z^{*}$ can be obtained by solving a simple linear program:
\begin{equation}
Z^{*} = \arg\max_{Z}\quad C^TZ,  \quad\quad
\text{s. t.} \quad\mathbf{1}^TZ=m^*.
\label{eq:inference_optimum}
\end{equation}
Note that the optimal solution to the problem in Eq.~(\ref{eq:inference_optimum}) are exactly those variables that correspond to the $m^*$ highest values of $C$. 

\section{Experimental Results}
\label{results}

To validate our proposed joint set learning approach, we perform experiments on the task of multi-label image classification. This is an appropriate application for our model as its output is expected to be in the form of a set (a set of labels in this particular case) with an unknown cardinality while the order of its elements (labels) in the output list does not have any meaning. Moreover, we assume that the labels are derived from an \iid-cluster process model. To make our work directly comparable to~\cite{rezatofighi2017deepsetnet}, we use the same two standard and popular benchmarks, the PASCAL VOC 2007 dataset~\cite{Everingham:2007:PASCAL-VOC} and the Microsoft Common Objects in Context (MS COCO) dataset~\cite{Lin:2014:COCO}.

\myparagraph{Implementation details.}
We follow the same experimental setup used in~\cite{rezatofighi2017deepsetnet,Wang_2016_CVPR}. Our model is built on the $16$-layers VGG network~\cite{Simonyan:2014:VGG}, pretrained on the 2012 ImageNet dataset. We adapt VGG for our purpose by modifying the last fully connected prediction layer to predict both cardinality and classification distributions according to the loss proposed in \Eq~(\ref{eq:map_dual}), \ie DC for cardinality and \textit{binary cross-entropy} for classification. We then fine-tune the entire network using the training set of these datasets with the same train/test split as in existing literature~\cite{rezatofighi2017deepsetnet,gong2013deep,Wang_2016_CVPR}. 

To train our network, which we call JDS in the following, we use stochastic gradient descent and set the weight decay to $\gamma = 5\cdot 10^{-4}$, with a momentum of $0.9$ and a dropout rate of $0.5$.  The learning rate is adjusted to gradually decrease after each epoch, starting from $0.001$. The network is trained for $60$ epochs for both datasets and the epoch with the lowest validation objective value is chosen for evaluation on the test set. The hyper-parameter $U$ is set to be $2.36$, adjusted on the validation set.   

To demonstrate that  joint learning is helpful to learn a better classifier (state distribution) as well as a better cardinality distribution, we perform an additional baseline experiment where we replace the negative binomial (NB) distribution used in~\cite{rezatofighi2017deepsetnet} with the Dirichlet-Categorical (DC) distribution from \Eq~(\ref{eq:DC}). Then, an independent cardinality distribution network is trained using the same network structure as the one used in~\cite{rezatofighi2017deepsetnet} while modifying the final fully connected layer to predict the cardinality using the DC distribution. 
We fine-tune the network on cardinality distribution, initialised with the network weights learned for the classification task of each of the reported datasets, \ie PASCAL VOC and MS COCO. To train the cardinality CNN, we use the exact same hyper-parameters and training strategy as described above.

\myparagraph{Evaluation protocol.}
% \subsubsection{Evaluation protocol}
We employ the common evaluation metrics for multi-label image classification also used in~\cite{gong2013deep,Wang_2016_CVPR,rezatofighi2017deepsetnet}. These include the average \textit{precision}, \textit{recall} and \textit{F1-score}\footnote{F1-score is calculated as the harmonic mean of precision and recall.} of the generated labels, calculated per-class (C-P,  C-R and C-F1) and overall (O-P,  O-R and O-F1). Since C-P,  C-R and C-F1 can be biased to the performance of the most frequent classes, we also report the average \textit{precision}, \textit{recall} and \textit{F1-score} of the generated labels per image/instance (I-P, I-R and I-F1). 

\begin{figure}[t]
	\includegraphics[width=0.85\linewidth]{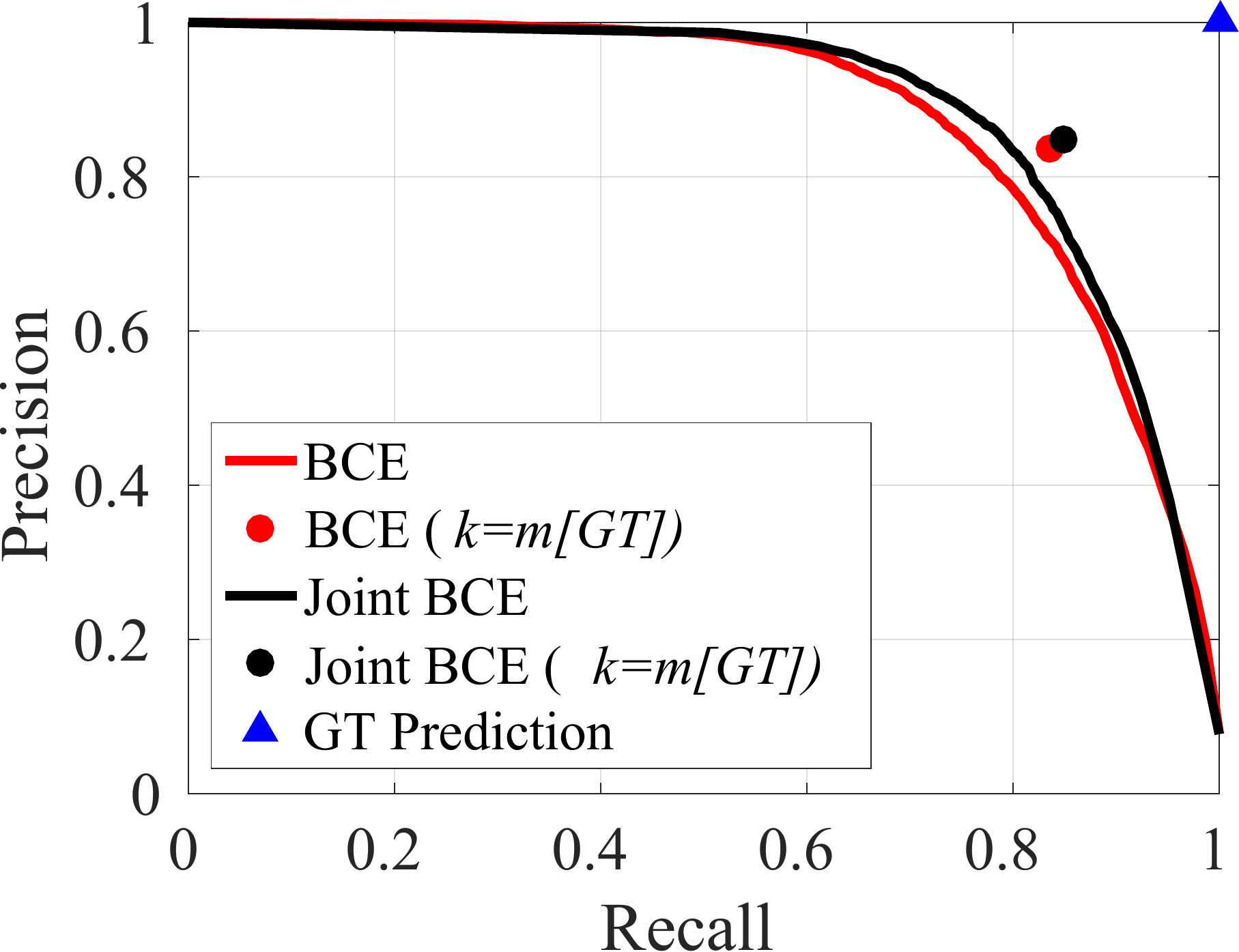}
	\caption{Precision/recall curves for the classification scores when the classifier is trained independently (red solid line) and when it is trained jointly with the cardinality term using our proposed joint approach (black solid line) on PASCAL VOC dataset. The circles represent the upper bound when ground truth cardinality is used for the evaluation of the corresponding  classifiers. The ground truth prediction is shown by a blue triangle.}\label{fig:curves-mlic}
\end{figure}

% Since precision and recall can be different for each approach and cannot be a proper representative metric, we rely on F1 score to rank approaches on the final task of label prediction. The perfect set prediction will have F1 score equal to 1, which is a single point in the top right corner of the precision and recall plot (see blue triangle in Fig.~\ref{fig:curves-mlic}). 
We rely on F1-score to rank approaches on the task of label prediction. A method with better performance has a precision/recall value that has a closer proximity to the perfect point shown by the blue triangle in Fig.~\ref{fig:curves-mlic}. % This property is well reflected in the F1 score. 
%  Precision is defined as the ratio of
%correctly predicted labels and total predicted labels, while
%recall is the ratio of correctly predicted labels and ground-truth labels. 
%In case no predictions (or ground truth) labels exist, \ie the denominator becomes zero, precision (or recall) is defined as $\%100$. 
To this end, for the classifiers such as BCE and Softmax, we find the optimal evaluation parameter $k=k^*$ that maximises the F1-score. For the deep set network (DS)~\cite{rezatofighi2017deepsetnet} and our proposed joint set network (JDS), prediction/recall is not dependent on the value of $k$. Rather,  one single value for precision, recall and F1-score is computed. 

% \subsubsection{PASCAL VOC 2007}
\myparagraph{PASCAL VOC 2007.} 
\newcommand{\colw}{0.43cm}
\newcommand{\colww}{0.52cm}
\newcommand{\lsh}{\!\!\!\!}
\begin{table*}[!h]
	\caption{Quantitative results for multi-label image classification on the PASCAL VOC dataset.}
%	\vspace{-1em}
	\begin{center}
		\begin{tabular}{lc||p{\colw}p{\colw}p{\colww}| p{\colw}p{\colw}p{\colww}|p{\colw}p{\colw}p{\colww} @{}}
			% 			\raisebox{-1.0ex}{Classifier}& \raisebox{-1.0ex}{/Metric} & \raisebox{-1.0ex}{C-P} & \raisebox{-1.0ex}{C-R} & \raisebox{-1.0ex}{C-F1} & \raisebox{-1.0ex}{O-P} & \raisebox{-1.0ex}{O-R} & \raisebox{-1.0ex}{O-F1}\\
			% 			&\raisebox{-0.5ex}{Evaluation}&&&&&&\\
			\lsh Classifier & Eval. & \scriptsize{C-P} & \scriptsize{C-R} & \scriptsize{C-F1} & 
			\scriptsize{O-P} & \scriptsize{O-R} & \scriptsize{O-F1} & \scriptsize{I-P} & \scriptsize{I-R} & \scriptsize{I-F1} \\
			\hline\hline % inserts single-line
			% Entering 1st row
			%			Upper Band&k=3&$64.1$&$74.4$&$68.9$&$75.1$&$78.3$&$76.7$\\ 
			%			\hline
			\lsh Softmax&k=$k^*$(1)&$88.2$&$65.4$&$75.1$ &$91.3$&$59.2$&$71.8$&$91.3$&$69.8$&$79.1$\\
			%			WARP&k=3&&&&&&\\
			\lsh BCE&k=$k^*$(1)&$88.7$&$58.6$&$70.5$&$92.2$&$59.8$&$72.5$&$92.2$&$70.1$&$79.6$\\
			
			\lsh DS (BCE-NB)~\cite{rezatofighi2017deepsetnet}&k=$m^*$ &$76.8$&$74.8$&$75.8$&$80.6$&$76.7$&$78.6$&$83.4$&$81.9$&$82.6$\\
			%			\lsh DS (BCE-NB)~\cite{}&k=$m^{GT}$ &$78.8$&$80.8$&$79.8$&$83.5$&$83.5$&$83.5$&$85.7$&$85.7$  &$85.7$\\
			\lsh DS (BCE-DC)&k=$m^*$&$77.1$ & $75.2$ &$76.2$ & $81.0$ &  $77.1$ & $79.0$ & $83.9$ & $82.1$ & $83.0$\\
			\hline
			\lsh\textbf{JDS (BCE-DC)}&k=$m^*$ &$83.5$&$74.4$&$\textbf{78.7}$&$85.5$&$77.9$&$\textbf{81.5}$&$87.6$&$82.8$&$\textbf{85.1}$\\
		\end{tabular}
	\end{center}
	\label{table:allvoc-multilabel}
%	\vspace{-.5em}
\end{table*}
We first test our approach on the Pascal Visual Object Classes benchmark~\cite{Everingham:2007:PASCAL-VOC}, which is one of the most widely used datasets for detection and classification. This dataset includes $9963$ images with a 50/50 split for training and test, where objects from $20$ pre-defined categories have been annotated by bounding boxes. Each image  contains between $1$ and $7$ unique objects. 

We first investigate if the joint learning improves the performance of cardinality and classifier. Fig.~\ref{fig:curves-mlic} shows the precision/recall curves for the classification scores when the classifier is trained solely using binary cross-entropy (BCE) loss (red solid line) and when it is trained using the same loss jointly with the cardinality term (Joint BCE). We also evaluate the precision/recall values when the ground truth cardinality $m[GT]$ is provided. The results confirm our claim that the joint learning indeed improves the classification performance. We also calculate the mean absolute error of the cardinality estimation when the cardinality term using the DC loss is learned jointly and independently as proposed in~\cite{rezatofighi2017deepsetnet}. The mean absolute cardinality error of our prediction on PASCAL VOC is $0.31\pm0.54$, while this error is $0.33\pm0.53$ when the cardinality is learned independently. 
We compare the performance of our proposed joint deep set network, \ie JDS (BCE-DC), with softmax and BCE classifiers with the best $k$ value as well as the deep set network~\cite{rezatofighi2017deepsetnet} when the classifier is binary cross entropy and the cardinality loss is negative binomial, \ie DS (BCE-NB). In addition, Table~\ref{table:allvoc-multilabel} reports the results for the deep set network when the cardinality loss is replaced by our proposed Dirichlet-Categorical loss, \ie (BCE-DC). The results show that we outperform the other approaches \wrt all three types of F1-scores. In addition, our joint formulation allows for a single training step to obtain the final model, while the deep set network learns two VGG networks to generate the output sets.

% % \fixmea{Report cardinality estimation error?}\fixmeh{mean absolute error = $0.3241$ and std absolute error  = $0.5235$ on the test set}

%  \begin{figure*}[t]
%  	\centering
%  	\includegraphics[width=.49\linewidth]{figs/PerClass_ROC_Curve_VOC.pdf}
%  	\includegraphics[width=.49\linewidth]{figs/Overal_ROC_Curve_VOC.pdf}
%  	\caption{Pascal VOC 2007.} 
%  	\label{fig:curvesVOC}
%  	\end{figure*}

\myparagraph{Microsoft COCO. } 
\begin{figure*}[tb]
	\includegraphics[width=1.03\linewidth,left]{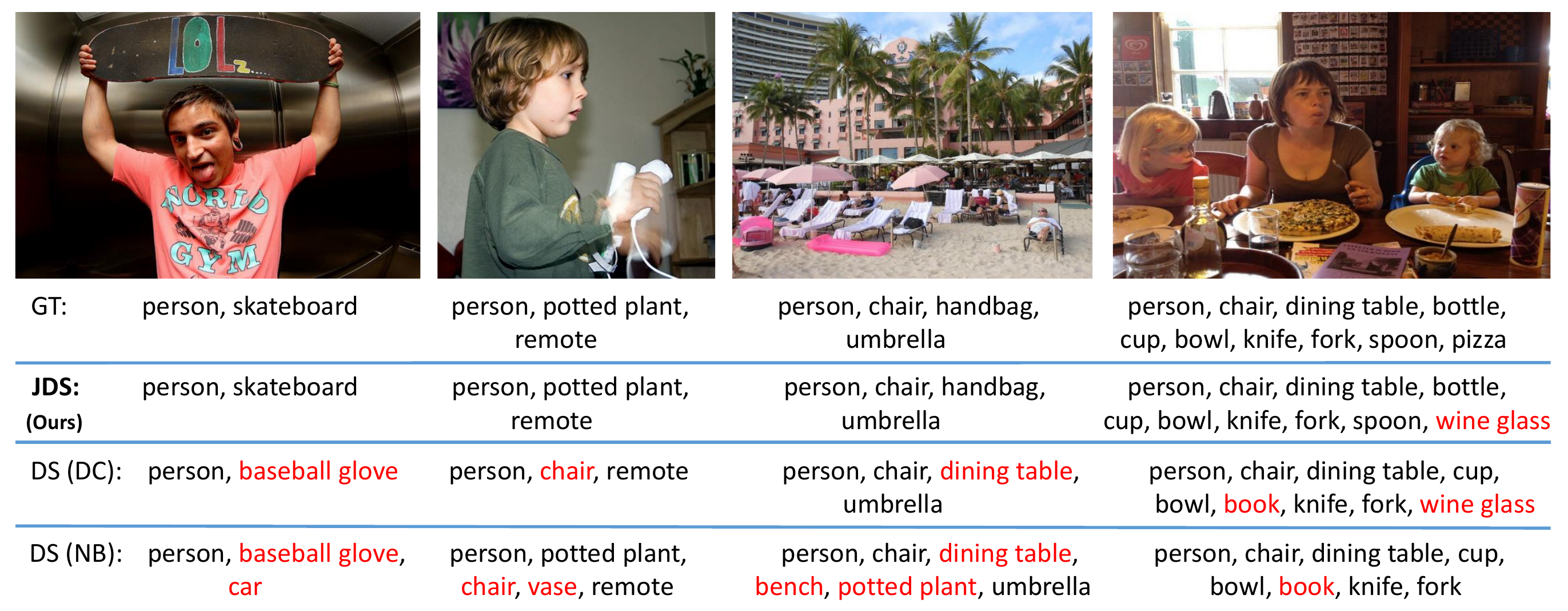}
	\caption{Qualitative comparison between our proposed joint deep set network (JDS) and the deep set networks with Dirichlet-Categorical (DS (DC)) and Negative Binomial (DS (NB)) as the cardinality loss. For each image, the ground truth tags and the predictions for our JDS and the two baselines are denoted below. {\textcolor{red}{False positives}} are highlighted in red. Our JDS approach reduces both cardinality and classification error.} 
%	\vspace{-1.0em}
	\label{fig:Results1}
\end{figure*}
\begin{table*}[!h]
	\caption{Quantitative results for multi-label image classification on the MS COCO dataset.}
%	\vspace{-1em}
	\begin{center}
		\begin{tabular}{lc||p{\colw}p{\colw}p{\colww}| p{\colw}p{\colw}p{\colww}|p{\colw}p{\colw}p{\colww} @{}}
			% 			\raisebox{-1.0ex}{Classifier}& \raisebox{-1.0ex}{/Metric} & \raisebox{-1.0ex}{C-P} & \raisebox{-1.0ex}{C-R} & \raisebox{-1.0ex}{C-F1} & \raisebox{-1.0ex}{O-P} & \raisebox{-1.0ex}{O-R} & \raisebox{-1.0ex}{O-F1}\\
			% 			&\raisebox{-0.5ex}{Evaluation}&&&&&&\\
			\lsh Classifier & Eval. & \scriptsize{C-P} & \scriptsize{C-R} & \scriptsize{C-F1} & 
			\scriptsize{O-P} & \scriptsize{O-R} & \scriptsize{O-F1} & \scriptsize{I-P} & \scriptsize{I-R} & \scriptsize{I-F1} \\
			\hline\hline % inserts single-line
			% Entering 1st row
			%			Upper Band&k=3&$64.1$&$74.4$&$68.9$&$75.1$&$78.3$&$76.7$\\ 
			%			\hline
			
			\lsh Softmax&k=$k^*$(3)&$58.6$&$57.6$&$58.1$ &$60.7$&$63.3$&$62.0$&$60.7$&$74.7$&$67.0$\\
			%			WARP&k=3&&&&&&\\
			\lsh BCE&k=$k^*$(3)&$56.2$&$60.1$&$58.1$&$61.6$&$64.2$&$62.9$&$61.6$&$75.3$&$67.8$\\
			\lsh CNN-RNN~\cite{Wang_2016_CVPR}&k=$k^*$(3)&$66.0$&$55.6$&$60.4$&$69.2$&$66.4$&$67.8$&$-$&$-$&$-$\\
			\lsh DS (Softmax-NB)~\cite{rezatofighi2017deepsetnet}&k=$m^*$ &$68.2$&$59.9$&$63.8$&$68.8$&$67.4$&$68.1$&$74.3$ & $72.6$ & $73.5$\\
			%			\textbf{DeepSetNet (WARP)}&Est. Card.&&&&&&\\
			\lsh DS (BCE-NB)~\cite{rezatofighi2017deepsetnet}&k=$m^*$ &$66.5$&$62.9$&$64.6$&$70.1$&$68.7$&$69.4$&$75.2$&$73.6$&$74.4$\\	
			%			WARP&k=3&&&&&&\\
			\lsh DS (BCE-DC)& k=$m^*$& $68.0$ & $61.7$ & $64.7$ & $72.4$ & $67.1$ & $69.6$& $76.0$ & $73.3$ & $74.6$\\
			\hline
			\lsh\textbf{JDS (BCE-DC)}&k=$m^*$ &$70.2$&$61.5$&$\textbf{65.5}$&$74.0$&$67.6$&$\textbf{70.7}$&$77.9$&$73.4$&$\textbf{75.6}$\\
			
			%			\hline
			%			\lsh\textbf{JDS (BCE-DC)}&k=$m^{GT}$ &&&&&&&&&\\
			%			\lsh\textbf{JDS (Rnk-DC)}&k=$m^*$&&&&&&&&&\\
		\end{tabular}
	\end{center}
	\label{table:allcoco-multilabel}
%	\vspace{-1.5em}
\end{table*}
The MS-COCO~\cite{Lin:2014:COCO} benchmark is another popular benchmark for image captioning, recognition, and segmentation. The dataset includes
$123$K  images, each labelled with per instance
segmentation masks of $80$ classes. The number of unique objects for each image  varies between $0$ and $18$. Around $700$ images in the training set do not contain any of the $80$ classes and there are only a handful of images that have more than $10$ tags. Most images contain between one and three labels.  We use $82783$
images with identical training and validation split as~\cite{rezatofighi2017deepsetnet}, and the remaining $40504$ images as test data. 
%We predict the cardinality of objects in the scene with a mean absolute error of $0.74\pm0.86$.
%%
%%\input{tables/allcoco-multilabel}
%
%
%\Fig~\ref{fig:curves-mlic}(b) shows a significant improvement of precision and recall and consequently the F1 score using our deep set network compared to the softmax and binary cross-entropy classifiers for all ranking values $k$. 

The classification results on this dataset are reported in Table~\ref{table:allcoco-multilabel}. The results once again show that our approach consistently outperforms our baselines and the other methods measured by F1-score. Due to this improvement, we achieve  state-of-the-art results on this dataset as well. 
Some examples of label prediction using our joint deep set network and comparison with other deep set networks are shown in \Fig~\ref{fig:Results1}. The results show that our joint learning can simultaneously reduce the cardinality and classification errors in these examples.  
%We also investigated failure cases where either the cardinality CNN or the classifier fails to make a correct prediction. We showcase some of these cases in Fig~\ref{fig:Results2}. We argue here that some of the failure cases are simply due to a missed ground truth annotation, such as the left-most example, but some are actually semantically correct \wrt the cardinality prediction, but are penalized during evaluation because a particular object category is not available in the dataset. This is best illustrated in the second example in \Fig~\ref{fig:Results2}. Here, our network correctly predicts the number of objects in the scene, which is two, however, the can does not belong to any of the 80 categories in the dataset and is thus not annotated. Similar situations also appear in other images further to the right.

\section{Conclusion}
We proposed a framework to jointly learn and predict a set's cardinality and state distributions by modelling both distributions using the same set of weights. This approach not only significantly reduces the number of learnable parameters, but also helps to model both distributions more accurately. We have demonstrated the effectiveness of this approach on multi-class image classification, outperforming previous state of the art on standard datasets.  
The main limitation of our framework is that we do not include the complexity of permutation invariance of sets in the learning step. Therefore, our method is only applicable to set problems that do not rely on permutation invariance during training, such as image tagging. In future, we plan to overcome this limitation by incorporating permutation variables during training procedure. Another potential avenue could be to exploit the Bayesian nature of the model to include uncertainty as opposed to relying on the MAP estimation alone. 

\myparagraph{Acknowledgments.}
This research was supported by the Australian Research Council through the Centre of Excellence in Robotic Vision, CE140100016, and through Laureate Fellowship FL130100102 to IDR.

\bibliographystyle{aaai}
\bibliography{reference/ref,reference/refs-short,reference/anton-ref}

\end{document}